\PassOptionsToPackage{table}{xcolor}
\documentclass[11pt]{article}
\usepackage{acl}

\usepackage{times}
\usepackage{latexsym}
\usepackage[T1]{fontenc}
\usepackage[utf8]{inputenc}
\usepackage{microtype}
\usepackage{inconsolata}
\usepackage{booktabs}
\usepackage{multirow}
\usepackage{graphicx}
\usepackage{verbatim}
\usepackage{amsmath}
\usepackage{hyperref}
\usepackage{url}
\usepackage{array}
\graphicspath{{figures/}}
\usepackage{xcolor} 
\usepackage[most]{tcolorbox} 
\definecolor{revblue}{RGB}{0,0,0}
\newcommand{\new}[1]{\textcolor{revblue}{#1}}
\newcommand{\newpar}[1]{{\color{revblue}#1}}
\definecolor{editblue}{RGB}{0,0,0}
\newcommand{\edit}[1]{\textcolor{editblue}{#1}}
\newcommand{\editpar}[1]{{\color{editblue}#1}}

\newtcolorbox{promptbox}[1][]{%
  colback=blue!4!white,
  colframe=blue!55!black,
  coltitle=white,
  fonttitle=\bfseries\small,
  title=#1,
  boxrule=0.5pt, arc=2pt,
  left=5pt, right=5pt, top=4pt, bottom=4pt,
  before skip=4pt, after skip=6pt,
  breakable,
  fontupper=\footnotesize
}
\title{Can One Adapted Model Do It All? \\ Fine-Tuning Strategy Selection for Customer Support LLMs\\\texorpdfstring{%
  \raisebox{-0.2\height}{\includegraphics[height=7mm]{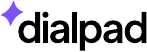}}%
}{}
}

\author{Md Tahmid Rahman Laskar, Xue-Yong Fu, Shashi Bhushan TN\\\{\texttt{tahmid.rahman,xue-yong,sbhushan}\}\texttt{@dialpad.com}\\{Dialpad Inc.}}

\begin{document}
\maketitle

\begin{abstract}
Production customer-support systems often require LLMs to support multiple skills, such as intent classification, question answering, summarization, or tool-use decisions. A central deployment question is whether these skills should be handled by separate task-specialist models or by a single model trained through multi-task training, sequential updates, or model merging. We study this question using \new{thirteen models spanning five families (Qwen3, Qwen3.5, Gemma-3, Llama-3.1, and Mistral) from 0.6B to 32B parameters} across eight customer-support datasets, spanning four public and four proprietary datasets with \edit{74k training and 8.6k evaluation samples}. Under a fixed training protocol, we train \new{more than 200 checkpoints and fine that}
\edit{multi-task full fine-tuning provides a practical operational default when a single checkpoint is required, with performance close to task specialists for most models.} 
Specialist models are strong on their target tasks but often degrade sharply off-task, making reliable routing important. Sequential Low-Rank Adaptation (LoRA)  preserves earlier skills better than sequential full fine-tuning, while merging a specialist with its base model improves off-task robustness with limited same-task loss for larger models. We conclude with practical guidelines for selecting fine-tuning strategies in real-world settings. 
\end{abstract}

\section{Introduction}
\label{sec:intro}

Production conversational assistants in customer support are expected to perform multiple tasks, such as identifying a customer's intent, answering knowledge-base questions, summarizing a conversation, or deciding whether to call an external tool \cite{zhang2025reic,laskar2023building,laskar2025ai,alkhouli2025confetti}. Real-world deployments favor cost-effective, smaller LLMs, which lack the general-purpose capability of larger ones and therefore must be adapted to target tasks via fine-tuning \cite{fu2024tiny}. This raises a practical design question: should practitioners train a separate model for each skill, or one model that performs all skills?

This choice has direct deployment implications. Separate specialist models can provide strong task-specific performance, but they require more storage, more endpoints, and reliable routing to the appropriate specialist model \cite{chen2023frugalgpt,ding2024hybrid,ongroutellm}. In contrast, models fine-tuned across multiple tasks can serve many tasks through a single checkpoint. However, updating one task may change the behavior of others. While other strategies can reduce some of these costs, they may also introduce different trade-offs. For instance, parameter-efficient fine-tuning techniques like LoRA \citep{hu2022lora} keep a shared base model and train small task-specific adapters that can reduce storage and update cost; they may fail to match the performance of full fine-tuning. Techniques like sequential fine-tuning \cite{fu2022effective} can be useful when new tasks are added over time, but they can degrade performance on earlier tasks. Post-hoc model merging methods \citep{wortsman2022model,ilharco2023editing,yu2024language} can combine trained checkpoints without additional fine-tuning, but their ability to preserve multiple customer-support skills remains unclear.

Despite these trade-offs, practitioners may choose among these strategies using informal rules rather than controlled evidence. Prior work has studied multi-task learning \citep{raffel2020t5,sanh2022multitask}, adapter-based fine-tuning \citep{hu2022lora,pfeiffer2021adapterfusion}, continual learning \citep{parisi2019continual,delange2021continual}, and model merging \citep{wortsman2022model,yadav2023ties} separately. However, these methods are rarely compared under the same training setup, model family, and evaluation protocol, particularly on real-world customer-support workloads that require multiple deployment-relevant skills. This makes it difficult to know whether performance differences come from the strategy itself or from changes in data, models, or evaluation.

To this end, we present a controlled comparison of different fine-tuning strategies for customer support use cases that covers: task-specific full fine-tuning, LoRA adaptation, multi-task fine-tuning, sequential fine-tuning, and model merging. 
Our experiments span \new{thirteen models across five families and parameter counts from 0.6B to 32B}, \new{yielding over 200 trained models.}
Our key insights are:

\edit{(i) Multi-task full fine-tuning offers a practical single-checkpoint default, although specialists can provide model-specific performance advantages;} 

(ii) Specialist models often perform well on their target task but degrade sharply on others, which matters when routing is imperfect;

(iii) Sequential full fine-tuning substantially hurts earlier skills, while LoRA can mitigate; 

\edit{(iv) Mixing a specialist with its base improves average off-task performance; for the evaluated models of at least 4B parameters, suite-average in-task losses are at most 2.6 points;}

Based on these findings, we provide practical recommendations for choosing a strategy based on task similarity, model size, and update frequency. 

\section{Related Work}
\label{sec:related}

\noindent\textbf{Fine-tuning and parameter-efficient adaptation.}
Full fine-tuning adapts all model parameters and can provide strong task-specific performance, but is computationally expensive and may reduce general capabilities \citep{devlin2019bert,raffel2020t5,xu2026parameter}. Parameter-efficient methods such as LoRA instead update a small set of parameters while keeping the base model frozen \citep{hu2022lora}. We compare both approaches in this work. 

\noindent\textbf{Multi-task learning and task specialization.}
Multi-task learning trains one model across tasks and can improve data efficiency and generalization \citep{raffel2020t5,sanh2022multitask}, although its effectiveness depends on task relatedness, data composition, and model capacity \citep{wei2022finetuned,ouyang2022training}. Task-specific fine-tuning may perform better in-task but generalize poorly to other tasks \citep{wang2024two}, creating a trade-off between specialization and maintaining a limited number of deployment endpoints.

\noindent\textbf{Sequential updates and catastrophic forgetting.}
Sequential adaptation supports tasks added over time but can cause catastrophic forgetting, in which later updates degrade earlier capabilities \citep{goodfellow2013empirical,parisi2019continual,delange2021continual}. This risk is especially relevant for models repeatedly updated for new customer-support tasks.

\noindent\textbf{Model merging.}
AdapterFusion \citep{pfeiffer2021adapterfusion} and LoRAHub \citep{huang2024lorahub} learn to combine task-specific adapters using supervision or few-shot examples. Post-hoc merging instead combines independently fine-tuned parameters without additional training \citep{wortsman2022model,ilharco2023editing}, but conflicting task updates can cause negative transfer \citep{yadav2023ties}. In this paper, we investigate post-hoc model merging that does not require any re-training. 

To our knowledge, these adaptation strategies have not been systematically compared under a unified protocol across customer-support tasks yet.



\section{Methodology}
\label{sec:setup} %
\subsection{Adaptation Strategies}

We compare several adaptation strategies for production customer-support LLMs under a fixed model training and evaluation protocol. 
Specifically, our methodology includes single-task and multi-task fine-tuning, parameter-efficient fine-tuning with LoRA, sequential fine-tuning for continual-task updates, and post-hoc model merging of the trained model(s) with the base model or between them. 
Below, we discuss various adaptation strategies that we study.  

\noindent \textbf{Single-task and multi-task adaptation.}
We compare task-specialist models with unified multi-task models. In the single-task setting, we train one model per task and evaluate it on its target task alongside other remaining tasks in the evaluation suite. This measures task specialization as well as generalization and robustness. In the multi-task setting, we train one model on a training set that 
covers all tasks within an evaluation suite. This produces one single checkpoint intended to serve multiple customer-support features.

\noindent \textbf{Full and parameter-efficient fine-tuning.}
We evaluate both full fine-tuning and parameter-efficient fine-tuning. Full fine-tuning updates all model parameters, whereas parameter-efficient fine-tuning uses LoRA adapters while keeping the base model fixed. We apply these strategies in both single and multi-task settings to compare target-task performance, as well as cross-task retention. 

\noindent \textbf{Sequential fine-tuning and continual adaptation.}
To simulate settings where new skills are added over time, we fine-tune models sequentially across tasks. We compare sequential full fine-tuning with sequential LoRA and evaluate the final checkpoint on all tasks. This setup allows us to measure catastrophic forgetting, i.e., whether adapting to later tasks degrades performance on earlier ones.

\noindent \textbf{Model merging.}
\editpar{We evaluate two distinct post-hoc methods: linear interpolation between each specialist and its original base model, and TIES merging \citep{yadav2023ties} of multiple independently trained specialists. The former tests recovery of off-task performance while retaining in-task gains; the latter tests whether independently learned skills can be combined without joint training.}


\subsection{Datasets}
\label{sec:datasets}
We evaluate different fine-tuning strategies across diverse customer-support datasets, covering tasks from public academic benchmarks and real-world proprietary benchmarks. We discuss these benchmarks below (we also provide the prompts for all tasks in Appendix \ref{app:prompts}).
\subsubsection{Public Academic Suite}
\label{ssec:public-suite}
The public suite comprises four tasks (total \edit{$56{,}803$ training and $6{,}607$ evaluation samples}):

  \noindent \textbf{Intent Classification (IC).} We use the Bitext customer support
  dataset for intent classification, which contains $26{,}872$ training\footnote{\url{https://github.com/bitext/customer-support-llm-chatbot-training-dataset}} samples and $818$ evaluation\footnote{\url{https://www.kaggle.com/datasets/scodepy/customer-support-intent-dataset}} samples. \edit{The dataset contains 27 intent labels grouped into 11 high-level categories, consistent with the domain labels in Appendix~\ref{app:prompts}.}  

  \noindent \textbf{Dialogue Summarization (DS).} Conversation summarization is an important feature in customer support \cite{laskar2023building,laskar2024query}. For this, we use the SAMSum \citep{gliwa2019samsum} dataset
  ($14{,}731$ train, $1{,}637$ eval).

  \noindent  \textbf{Question Answering (QA).} We use WixQA \citep{cohen2025wixqa} for retrieval-grounded QA over a product knowledge base, with retrieved articles provided as context and answers required to cite evidence. The \emph{simulated} split is used for training ($200$) and the \emph{expert-written} split for evaluation ($200$). \new{Note that WixQA is simultaneously low-resource {and} distribution-shifted: the training split is simulated while the evaluation split is expert-written, and the two differ in document length, formatting, and citation style. This makes it a challenging benchmark for evaluation.}
 
  \noindent \textbf{Chatbot Response (CS).} We use the When2Call \citep{nvidia2024when2call}
  ($15{,}000$ train; $3{,}952$ eval) 
  dataset to investigate whether the model can correctly decide when to refrain from generating hallucinated answers, ask a follow-up question, or call a tool while responding to a user in a conversation. \edit{We evaluate generated responses using ROUGE-1, which adapts the original benchmark protocol and measures lexical overlap rather than decision accuracy.}

\begin{table*}[t]
\centering\small
\setlength{\tabcolsep}{3pt}
{%
\begin{tabular}{llrrrrrrrrrr}
\toprule
& & \multicolumn{5}{c}{Public suite} & \multicolumn{5}{c}{Internal suite} \\
Base & Method & IC & QA & DS & CS & Average & AI & CO & MS & CP & Average \\
\midrule
Qwen3-0.6B & Zero-shot & 38.8 & 28.8 & 36.2 & 23.0 & 31.7 & 31.2 & 2.9 & 33.3 & 11.0 & 19.6 \\
 & ST--Full & 91.6 & 37.2 & 47.8 & 40.3 & 54.2 & 54.0 & 73.3 & 62.7 & 47.0 & 59.3 \\
 & ST--LoRA & 94.7 & 40.9 & 49.3 & 32.5 & 54.4 & 54.4 & 42.9 & 61.8 & 48.0 & 51.8 \\
 & MT--Full & 91.6 & 34.8 & 47.3 & 40.1 & 53.5 & 57.1 & 67.6 & 64.2 & 52.5 & 60.4 \\
 & MT--LoRA & 96.3 & 40.2 & 36.0 & 31.9 & 51.1 & 51.2 & 58.1 & 60.1 & 50.5 & 55.0 \\
\midrule
Qwen3-1.7B & Zero-shot & 67.1 & 44.8 & 36.6 & 25.9 & 43.6 & 40.9 & 14.6 & 46.7 & 25.0 & 31.8 \\
 & ST--Full & 99.0 & 37.8 & 48.4 & 41.5 & 56.7 & 61.1 & 76.2 & 64.8 & 56.0 & \edit{64.5} \\
 & ST--LoRA & 97.6 & 44.6 & 49.5 & 38.1 & 57.4 & 57.9 & 42.2 & 63.3 & 53.5 & 54.2 \\
 & MT--Full & 99.0 & 38.1 & 50.8 & 39.0 & 56.7 & 55.6 & 54.6 & 65.2 & 56.5 & 58.0 \\
 & MT--LoRA & 95.2 & 36.8 & 51.3 & 36.3 & 54.9 & 57.7 & 41.3 & 63.7 & 51.5 & 53.5 \\
\midrule
Qwen3.5-2B & Zero-shot & 77.3 & 43.2 & 37.3 & 27.7 & 46.4 & 37.7 & 29.2 & 44.8 & 27.5 & 34.8 \\
 & ST--Full & 99.5 & 37.4 & 51.6 & 41.5 & 57.5 & 62.5 & 83.2 & 65.4 & 57.0 & 67.0 \\
 & ST--LoRA & 99.5 & 39.1 & 51.8 & 40.8 & 57.8 & 62.4 & 77.8 & 65.2 & 57.5 & 65.7 \\
 & MT--Full & 99.6 & 33.3 & 51.1 & 39.6 & 55.9 & 61.5 & 85.4 & 64.9 & 60.0 & 68.0 \\
 & MT--LoRA & 98.4 & 39.1 & 51.0 & 39.0 & 56.9 & 61.9 & 80.6 & 64.7 & 61.0 & 67.1 \\
\midrule
Qwen3-4B & Zero-shot & 88.0 & 47.5 & 34.7 & 24.6 & 48.7 & 40.0 & 39.4 & 47.7 & 40.0 & 41.8 \\
 & ST--Full & 99.8 & 37.3 & 52.3 & 42.9 & 58.0 & 64.6 & 85.7 & 67.1 & 57.5 & 68.7 \\
 & ST--LoRA & 98.2 & 44.7 & 53.1 & 40.6 & 59.1 & 64.0 & 79.7 & 67.0 & 60.5 & 67.8 \\
 & MT--Full & 98.9 & 36.8 & 52.2 & 36.2 & 56.0 & 63.7 & 79.7 & 66.7 & 57.0 & 66.8 \\
 & MT--LoRA & 98.2 & 40.9 & 52.7 & 36.9 & 57.2 & 64.1 & 74.6 & 67.1 & 58.0 & 65.9 \\
\midrule
Qwen3.5-4B & Zero-shot & 88.0 & 45.2 & 36.5 & 27.3 & 49.3 & 43.8 & 38.4 & 52.0 & 34.5 & 42.2 \\
 & ST--Full & 99.6 & 40.5 & 52.7 & 42.6 & 58.9 & 64.3 & 84.4 & 66.5 & 60.5 & 68.9 \\
 & ST--LoRA & 99.4 & 43.5 & 53.1 & 45.2 & 60.3 & 64.3 & 80.0 & 67.2 & 60.0 & 67.9 \\
 & MT--Full & 99.5 & 34.8 & 51.5 & 41.4 & 56.8 & 63.1 & 86.0 & 66.3 & 58.0 & 68.3 \\
 & MT--LoRA & 99.1 & 38.5 & 52.2 & 43.7 & 58.4 & 64.3 & 84.1 & 66.9 & 58.0 & 68.4 \\
\midrule
Qwen3-8B & Zero-shot & 89.7 & 49.0 & 38.2 & 29.4 & 51.6 & 48.1 & 37.5 & 51.6 & 41.5 & 44.7 \\
 & ST--Full & 99.5 & 39.6 & 53.0 & 42.3 & 58.6 & 64.9 & 84.1 & 67.6 & 60.0 & 69.1 \\
 & ST--LoRA & 97.8 & 45.8 & 53.5 & 37.8 & 58.7 & 61.8 & 71.7 & 67.0 & 59.5 & 65.0 \\
 & MT--Full & 97.7 & 39.8 & 52.6 & 42.1 & 58.1 & 64.7 & 78.4 & 67.1 & 59.5 & 67.5 \\
 & MT--LoRA & 98.2 & 43.3 & 53.4 & 37.1 & 58.0 & 62.5 & 68.9 & 66.6 & 55.5 & 63.4 \\
\midrule
Qwen3.5-9B & Zero-shot & 92.7 & 47.8 & 37.0 & 30.2 & 51.9 & 49.2 & 49.8 & 52.4 & 44.0 & 48.9 \\
 & ST--Full & 99.3 & 39.4 & 53.0 & 43.0 & 58.7 & 61.3 & 84.4 & 67.1 & 57.5 & 67.6 \\
 & ST--LoRA & 99.5 & 42.1 & 52.8 & 45.1 & 59.9 & 65.0 & 83.2 & 67.8 & 58.5 & 68.6 \\
 & MT--Full & 99.1 & 38.7 & 52.8 & 42.4 & 58.3 & 64.1 & 85.7 & 67.0 & 61.5 & 69.6 \\
 & MT--LoRA & 99.0 & 41.4 & 52.8 & 43.4 & 59.2 & 64.8 & 85.4 & 67.9 & 61.0 & 69.8 \\
\bottomrule
\end{tabular}}
\caption{Same-task performance under single-task (ST) and multi-task (MT) fine-tuning, full-parameter and LoRA, plus the zero-shot base per model. ST models are scored on the task they were trained for.}
\label{tab:mt-vs-st}
\end{table*}
\subsubsection{Proprietary Internal Suite}
\label{ssec:internal-suite}
The internal suite also comprises four tasks, which consist of real-world customer support
transcripts ($17{,}791$ training and $2{,}086$ evaluation examples in
total) 
collected from Dialpad\footnote{\url{https://www.dialpad.com/}}. The tasks are described below.

  \noindent \textbf{Action Items (AI).} This task focuses on producing a set of actionable follow-up tasks from a call transcript that should be completed after the call ends (5,755 train and 646 eval). 

  \noindent \textbf{Meeting Summarization (MS).} This task requires generating a concise summary of a business meeting according to specific constraints, such as the desired length (long, medium, or short) or format (for example, bullet points). It contains $7{,}829$ training and $925$ evaluation samples.

  \noindent \textbf{Call Purpose (CP).} This task requires the classification of the main purpose of a call into one of the predefined categories. For the call purpose, the model is asked to provide the call purpose category with an explanation ($1{,}596$ train, $200$ eval).

  \noindent \textbf{Call Outcome (CO).} This task involves identifying the primary outcome of a conversation by assigning it to one of the predefined call outcome categories without any additional text ($2{,}611$ train, $315$ eval).

\subsection{Models}
We evaluate thirteen models from five families, spanning 0.6B--32B parameters: Qwen3-0.6B, 1.7B, 4B, 8B, 14B, and 32B \citep{yang2025qwen3}; Qwen3.5-2B, 4B, and 9B\footnote{\url{https://huggingface.co/Qwen/Qwen3.5-9B}}; Gemma-3-1B and 4B \citep{gemmateam2025gemma3}; Llama-3.1-8B \citep{grattafiori2024llama3}; and Mistral-7B-Instruct-v0.3 \citep{jiang2023mistral7b}. The Qwen3 models are text-only, whereas Qwen3.5 is multimodal; the matched 4B variants help separate family effects from model size. All models were downloaded from Hugging Face \citep{wolf2019huggingface}. The hyperparameters that we use for training and inference, alongside implementation details, are given in Appendix \ref{hyperparameters}.  

\section{Results and Discussion}
\label{sec:results}
In this section, we present our experimental results. For brevity, we report the classification tasks: Intent Classification (IC), Call Outcome (CO), and Call Purpose (CP) using exact match, and the text generation tasks: Dialogue Summarization (DS), Meeting Summarization (MS), Action Items (AI), Question Answering (QA), and Chatbot Response (\edit{CS}) using ROUGE-1 \citep{lin2004rouge}. We additionally evaluate the generation tasks using BERTScore \cite{zhang2020bertscore} and an LLM judge.

\subsection{Main Results}
\label{sec:mainresults}

Table~\ref{tab:mt-vs-st} compares zero-shot, single-task (ST), and multi-task (MT) adaptation for seven representative models (Table \ref{tab:allmodels} in Appendix~\ref{app:allmodels} reports the remaining six). 
{We observe that ST-Full and MT-Full perform similarly for most models. More specifically, ST-Full leads on the public suite by 1.2 points on average and at most 2.8 points. \edit{On the internal suite, twelve of the thirteen models have gaps within $\pm2.5$ points; Qwen3-1.7B is the exception, with ST-Full averaging 64.5 versus 58.0 for MT-Full, a gap of approximately 6.5 points.}}
{MT-LoRA is generally close to MT-Full but trails on the internal suite by 5.4 points for Qwen3-0.6B, 4.5 for Qwen3-1.7B, and 4.1 for Qwen3-8B. Thus, its limitations are model-specific rather than governed by a consistent size threshold.}

Our additional analysis on the internal suite's call outcome task reveals that fine-tuned models almost always produce valid call-outcome labels, indicating that exact-match differences across models primarily reflect incorrect label selection rather than formatting. For example, Qwen3-1.7B ST-Full and MT-Full have similar validity rates (96.8\% vs.\ 96.2\%) despite differing substantially in accuracy (76.2 vs.\ 54.6).

\begin{figure}[t!]
    \centering
    \includegraphics[width=1\linewidth]{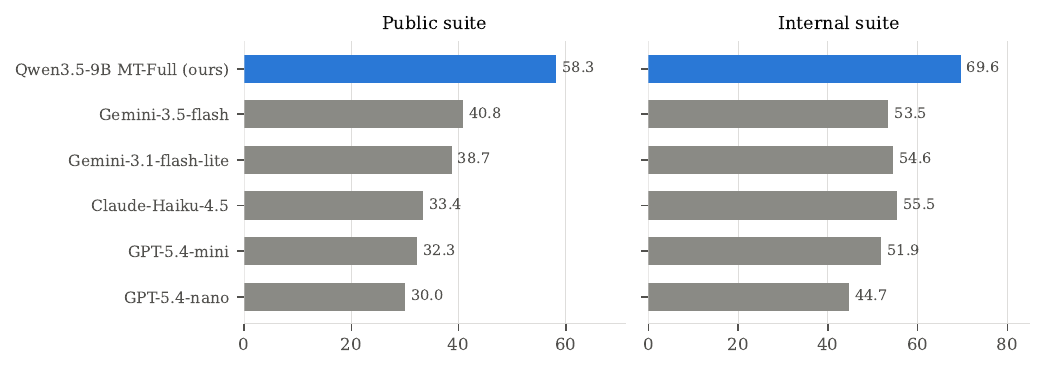}
    \caption{\small{\edit{Comparison between the evaluated closed zero-shot models and Qwen3.5-9B MT-Full.}}}
    \label{fig:zs_ft_9b}

\end{figure}

While comparing MT-LoRA and MT-Full, we find that MT-Full outperforms MT-LoRA across most models. Therefore, we further compare Qwen3.5-9B MT-Full with some proprietary zero-shot baselines on both suites in  Figure~\ref{fig:zs_ft_9b}. We observe that the Qwen3.5-9B MT-Full model consistently outperforms the zero-shot models. 

\begin{tcolorbox}[
colback=blue!5,
colframe=blue!20,
boxrule=0.4pt,
arc=1mm,
left=1mm,
right=1mm,
top=1mm,
bottom=1mm
]
\noindent\new{\textbf{Takeaway 1.}}
\edit{\emph{MT-Full is a practical default when one checkpoint is required: it performs close to task specialists for most models and avoids routing among specialist checkpoints. Specialist models remain appropriate when target-suite validation demonstrates a meaningful advantage.}}
\end{tcolorbox}

\begin{figure}[t]
    \centering
    \includegraphics[width=1\linewidth]{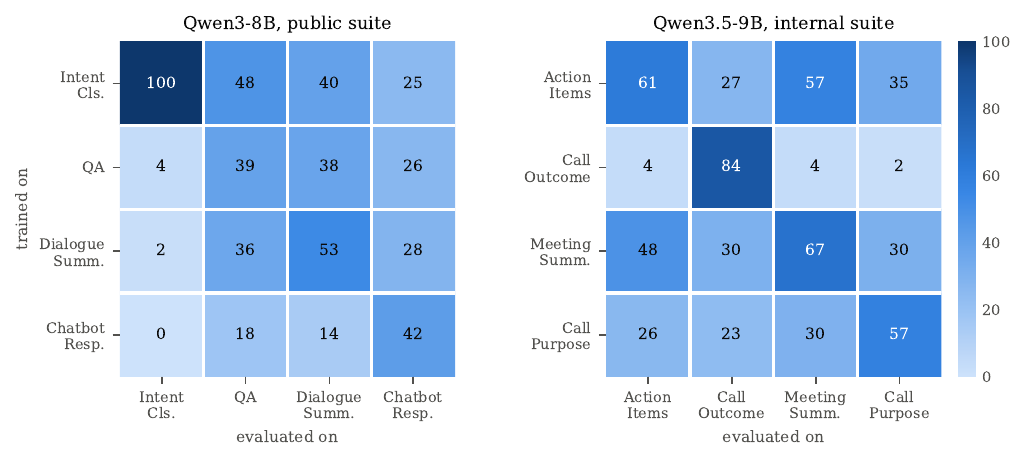}
    \caption{\small{Cross-task performance of single-task models: Qwen3-8B on the public suite and Qwen3.5-9B on the internal suite. Rows are the task a specialist was trained on, and columns are the tasks on which it is evaluated.}}
    \label{fig:cross-task}

\end{figure}

\begin{table}[t]
\centering\tiny
\setlength{\tabcolsep}{4pt}
\begin{tabular}{lcccccccc}
\toprule
& \multicolumn{4}{c}{\textbf{Qwen3-8B}} & \multicolumn{4}{c}{\textbf{Qwen3.5-9B}} \\
\cmidrule(lr){2-5} \cmidrule(lr){6-9}
\textbf{Method} & \textbf{IC} & \textbf{QA} & \textbf{DS} & \textbf{CS} & \textbf{AI} & \textbf{CO} & \textbf{MS} & \textbf{CP} \\
\midrule
ST Full & 99.5 & 39.6 & 53.0 & 42.3 & 61.3 & 84.4 & 67.1 & 57.5 \\
ST LoRA & 97.8 & 45.8 & 53.5 & 37.8 & 65.0 & 83.2 & 67.8 & 58.5 \\
Sequential Full & 25.9 & 21.0 & 48.4 & 42.4 & 46.1 & 70.5 & 52.9 & 59.5 \\
Sequential LoRA & 78.6 & 30.9 & 47.4 & 39.6 & 57.8 & 76.2 & 64.3 & 60.0 \\
\bottomrule
\end{tabular}
\vspace{-3mm}
\caption{\small{Sequential fine-tuning: IC $\rightarrow$ QA $\rightarrow$ DS $\rightarrow$ CS
for Qwen3-8B on the public suite, and AI $\rightarrow$ CO $\rightarrow$ MS $\rightarrow$ CP for
Qwen3.5-9B on the internal suite.}}
\label{tab:sequential}
\end{table}

\subsection{Cross-Task Robustness} \label{sec:cf}

In production, imperfect routing can send a specialist model a request meant for another model. Figure~\ref{fig:cross-task} shows that top ST specialists degrade sharply off-task. On the public suite, the Qwen3-8B chatbot-response specialist scores 42.3 on its own task but drops to 18.1 on QA and 14.4 on dialogue summarization. On the internal suite, the action-items and meeting-summary specialists score 57.1 and 48.3 on each other's tasks. Nonetheless, the label-producing call-outcome specialist performs very poorly on generation tasks (e.g., scores below 5.0 across generation tasks). 

\newpar{These findings suggest that transfer is stronger between tasks with similar outputs.  Specialists should therefore be deployed only with reliable routing or across tasks with compatible output requirements.}

\begin{tcolorbox}[
    colback=blue!5,
    colframe=blue!20,
    boxrule=0.4pt,
    arc=1mm,
    left=1mm,
    right=1mm,
    top=1mm,
    bottom=1mm
]
\noindent\textbf{Takeaway 2.}
\emph{Specialists are strong in-task but can fail badly off-task. Do not
deploy them unless routing is reliable.}
\end{tcolorbox}

\subsection{Sequential Fine-Tuning}
\label{sec:sequential}

To simulate incremental skill addition, we train Qwen3-8B in the order IC $\rightarrow$ QA $\rightarrow$ DS $\rightarrow$ CS and Qwen3.5-9B in the order AI $\rightarrow$ CO $\rightarrow$ MS $\rightarrow$ CP. Table~\ref{tab:sequential} reports the final checkpoint after all four stages.
\newpar{{Sequential full fine-tuning} causes substantial forgetting for ST-Full; intent classification falls from 99.5 to 25.9 and action items from 61.3 to 46.1. {Sequential LoRA} preserves these tasks better, retaining 78.6 and 57.8, respectively, although it trails the strongest non-sequential model.

 We further analyze the results by reversing the order and observe that earlier tasks suffer the most. Under reversed sequential full fine-tuning, the now-last intent-classification task retains 98.9, while the now-first chatbot-response task falls to 27.4 from \edit{the ST-Full specialist score of 42.3}. Sequential LoRA again retains the first internal task better than sequential full fine-tuning 
 (in the reverse order: 53.0 vs.\ 39.0). \edit{In the tested task orders, sequential LoRA generally preserves earlier skills better. Placing a critical task last can help protect its performance, although it can reduce retention of other tasks.}}

\begin{figure*}[t]
\centering
\includegraphics[width=0.9\textwidth]{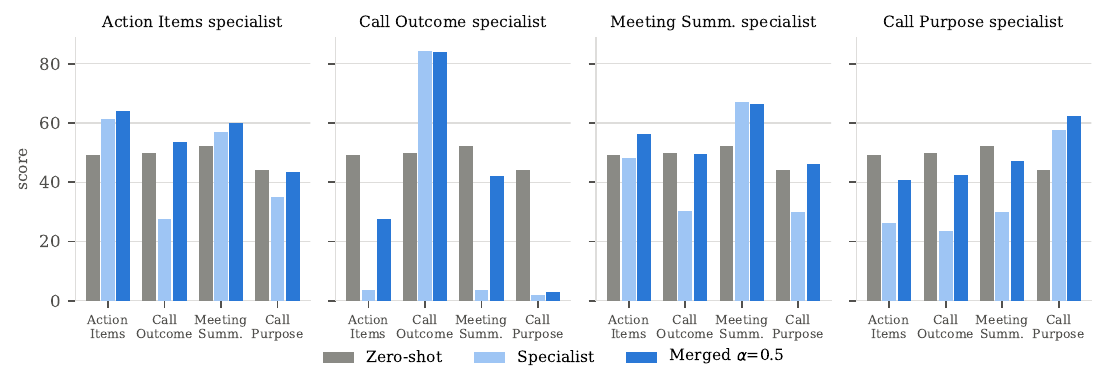}
\caption{\small{Cross-task results for Qwen3.5-9B on the internal suite after merging a specialist with its base at $\alpha=0.5$, compared with the unmerged specialist and the zero-shot base.}}
\label{fig:cross-merge}

\end{figure*}
 
\begin{table}[t]
\centering\tiny
\setlength{\tabcolsep}{5pt}
\begin{tabular}{lcccc}
\toprule
& \multicolumn{2}{c}{\textbf{Public Suite}} & \multicolumn{2}{c}{\textbf{Internal Suite}} \\
\cmidrule(lr){2-3} \cmidrule(lr){4-5}
\textbf{Base Model} & \textbf{In-task $\Delta$} & \textbf{Off-task $\Delta$} & \textbf{In-task $\Delta$} & \textbf{Off-task $\Delta$} \\
\midrule
Qwen3-0.6B & $-0.8$ & $+4.3$ & $-9.6$ & $+2.0$ \\
Qwen3-1.7B & $-1.6$ & $+4.0$ & $-6.4$ & $+7.0$ \\
Qwen3-4B & $+1.0$ & $+14.8$ & $-1.1$ & $+9.1$ \\
Qwen3-8B & $-0.2$ & $+6.9$ & $-2.6$ & $+10.9$ \\
Qwen3.5-9B & $+1.0$ & $+13.1$ & $+1.7$ & $+16.3$ \\
\bottomrule
\end{tabular}
\vspace{-2mm}
\caption{\small{Mean effect of merging each specialist with its base model
($\alpha=0.5$) across public and internal suites. }}
\label{tab:merge-net}
\end{table}

\begin{figure}[t!]
    \centering
    \includegraphics[width=0.8\linewidth]{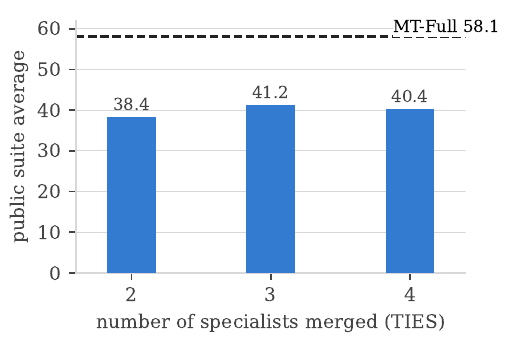}
    \caption{\small{TIES merging of $N$ public-suite specialists compared with multi-task fine-tuning.}}
    \label{fig:ties_zs_mt}
\end{figure}

\begin{tcolorbox}[
    colback=blue!5,
    colframe=blue!20,
    boxrule=0.4pt,
    arc=1mm,
    left=1mm,
    right=1mm,
    top=1mm,
    bottom=1mm
]
\noindent\textbf{Takeaway 3.}
\emph{Sequential LoRA is safer than sequential full fine-tuning, but retraining across multiple tasks or adding a new specialist is preferable when feasible.}
\end{tcolorbox}

 
\subsection{Model Merging}
\label{sec:merge}

We evaluate two post-hoc strategies: merging a specialist with its base model and merging multiple specialists. To merge a specialist with its base, for each ST checkpoint, we compute
$\theta_{\text{merged}}=\alpha\theta_{\text{FT}}+(1-\alpha)\theta_{\text{base}}$,
\edit{where $\alpha\in[0,1]$ and larger $\alpha$ favors the specialist. We primarily use $\alpha=0.5$ for consistency across configurations.} Table~\ref{tab:merge-net} shows that this approach consistently improves off-task robustness. \edit{For the evaluated models $\geq$4B, these gains come with at most a 2.6-point loss in suite-average in-task performance,} while Qwen3.5-9B improves both same- and off-task performance. Figure~\ref{fig:cross-merge} provides its task-level results.
Table~\ref{tab:alpha-ops} further illustrates the trade-off between specialization and robustness. As $\alpha$ increases from 0.3 to 0.7, off-task performance decreases on both suites, while in-task performance increases. Thus, $\alpha=0.3$ favors robustness and $\alpha=0.7$ favors specialization. 


\begin{table}[t]
\centering\scriptsize
\setlength{\tabcolsep}{5pt}
\begin{tabular}{lcccc}
\toprule
& \multicolumn{2}{c}{\textbf{Public}} & \multicolumn{2}{c}{\textbf{Internal}} \\
\cmidrule(lr){2-3}\cmidrule(lr){4-5}
$\alpha$ & \textbf{In-task} & \textbf{Off-task} & \textbf{In-task} & \textbf{Off-task} \\
\midrule
0.3 & 49.7 & 29.6 & 57.5 & 40.1 \\
0.5 & 57.1 & 27.5 & 62.3 & 37.3 \\
0.7 & 58.1 & 23.3 & 64.9 & 32.9 \\
\bottomrule
\end{tabular}
\vspace{-2mm}
\caption{\small{\new{Merge-with-base results for different $\alpha$ values averaged over the five
core models of different sizes (Qwen3 0.6B, 1.7B, 4B, 8B and Qwen3.5-9B).}}}
\label{tab:alpha-ops}
\end{table}

Merging multiple specialists does not match joint MT fine-tuning and fails to improve average score as more specialists (e.g., from 3 to 4) are added (Figure~\ref{fig:ties_zs_mt}). 

\begin{tcolorbox}[
    colback=blue!5,
    colframe=blue!20,
    boxrule=0.4pt,
    arc=1mm,
    left=1mm,
    right=1mm,
    top=1mm,
    bottom=1mm
]
\noindent \textbf{Takeaway 4.}
\emph{Merging a specialist with its base is a cheap robustness fix, while merging several specialists together cannot substitute MT-Full.}
\end{tcolorbox}
\subsection{\new{Beyond Lexical Overlap}}
\label{sec:semantic}

\newpar{We choose ROUGE-1 as a deterministic metric under which all strategies could be compared identically since our goal is the \emph{relative} effect of the adaptation strategy, not state-of-the-art result. \edit{We additionally report semantic evaluation for Qwen3-8B using BERTScore} \citep{zhang2020bertscore} and a Gemini-3-Flash judge that rates responses from 1--5.}
\newpar{\edit{The reported LLM-judge and BERTScore results in Table~\ref{tab:semantic} favor fine-tuned strategies over zero-shot and show small differences between ST-Full and MT-Full: BERTScore favors MT-Full on the public suite and ST-Full marginally on the internal suite.}}

\begin{table}[t]
\centering\tiny
\setlength{\tabcolsep}{4pt}
\begin{tabular}{lcccccc}
\toprule
& \multicolumn{2}{c}{\textbf{LLM judge}} & \multicolumn{2}{c}{\textbf{BERTScore}} & \multicolumn{2}{c}{\textbf{ROUGE-1}} \\
\cmidrule(lr){2-3}\cmidrule(lr){4-5}\cmidrule(lr){6-7}
\textbf{Strategy} & Public & Internal & Public & Internal & Public & Internal \\
\midrule
Zero--shot & 3.92 & 3.51 & 86.2 & 87.6 & 38.8 & 45.8 \\
ST--Full & 4.11 & 4.38 & 89.9 & 93.3 & 45.0 & 66.0 \\
MT--Full & 4.14 & 4.39 & 90.8 & 93.1 & 44.9 & 64.3 \\
MT--LoRA & 4.15 & 4.32 & 87.8 & 91.3 & 44.6 & 62.3 \\
\bottomrule
\end{tabular}
\vspace{-2mm}
\caption{\small{Average scores over generation tasks for Qwen3-8B.}}
\label{tab:semantic}
\end{table}

\subsection{Deployment Considerations}
\label{sec:deployment}

Post-training strategy selection should reflect endpoint limits, update frequency, and routing reliability. \edit{MT-Full is a practical choice when all task data are available and a single checkpoint is preferred, since it achieves performance close to task-specific models for most evaluated models. The larger internal-suite specialist advantage for Qwen3-1.7B illustrates the need for model-specific validation.} Specialists are appropriate only when they offer a substantial advantage and routing is reliable. When MT training is infeasible, sequential LoRA supports task updates with less forgetting, while merging specialists with their base improves off-task robustness. Appendix~\ref{sec:cost} further reports training times across model sizes.

\section{Conclusion}
\label{sec:conclusion}

We conducted a controlled comparison of post-training strategies across thirteen customer-support LLMs from five families and 0.6B--32B parameters. \edit{Single-task and multi-task full fine-tuning perform similarly for most models, although Qwen3-1.7B shows a larger internal-suite advantage for single-task training. No consistent size or capability threshold determines which is better.} More substantial differences emerge beyond same-task performance: specialists can fail sharply off-task, sequential full fine-tuning causes considerable forgetting, and LoRA better preserves earlier skills. Merging a specialist with its base improves robustness, whereas merging multiple specialists does not reliably replace joint training. Overall, post-training strategies should be evaluated through cross-task behavior and skill retention, rather than model scale or in-task performance alone.

\section*{Limitations}
\label{sec:limitations}

\paragraph{\new{Model coverage.}} \new{The study covers thirteen bases across five families (Qwen3, Qwen3.5, Gemma-3, Llama-3.1, Mistral) from 0.6B to 32B. Coverage is nonetheless uneven: the full strategy grid, including LoRA and merging variants, is run for a seven-model core, while the remaining models contribute single-task and multi-task full fine-tuning only. All bases are instruction-tuned; we do not test non-instruct checkpoints or models above 32B.}

\paragraph{Single language.} All training and evaluation datasets \new{in both suites are English-only, so all conclusions in this paper are English-only conclusions; whether the ST-versus-MT balance, the off-task collapse of specialists, or the forgetting behaviour of sequential updates hold under multilingual customer-support conversations is untested.} 


\paragraph{$\alpha$ choice.} Main-text merge-with-base tables use $\alpha=0.5$ for evaluation consistency \edit{across the reported merge configurations}. \new{Section~\ref{sec:merge} additionally reports $\alpha\in\{0.3,0.7\}$ as deployment operating points, and Appendix~\ref{app:alpha} gives the full sweep.}  

\paragraph{\new{Metric scope.}} \new{We choose ROUGE-1 as a deterministic metric for generation tasks. This is done so that all strategies can be compared identically since our goal is the \emph{relative} effect of the adaptation strategy, not the state-of-the-art result. Nonetheless, generation tasks are additionally evaluated by BERTScore and an LLM judge (\S\ref{sec:semantic}).}

\paragraph{Reproducibility.}{Half of this study runs on public data and is independently reproducible. We release the training and inference configurations for every run, the exact task prompts, the preprocessing scripts that build the four public datasets from their published sources, and the training and evaluation code, including the scoring implementations. The four internal datasets consist of PII-redacted enterprise call transcripts and cannot be released; the internal results are therefore reported as supporting production evidence rather than as an independently verifiable benchmark.}


\section*{Broader Impact and Ethics Statement}
\label{sec:ethics}

\noindent \textbf{1.} Helping practitioners pick the right strategy reduces wasted training compute and the energy footprint associated with it. The merge-with-base finding suggests many single-task fine-tunings can be made cross-task robust with no additional training, leading to direct compute savings. 

\noindent \textbf{2.} Our experiments use real customer-call transcripts, PII-redacted, in the proprietary internal suite. We do not release the proprietary data. 

\bibliography{custom}

\appendix

\section{Task Prompts}
\label{app:prompts}

We list the instruction templates used for each task. Placeholders in
square brackets (e.g., \texttt{[Categories]}, \texttt{[Transcript]}) and
braces (e.g., \texttt{\{Length Type\}}) denote task-specific content that is
filled in at runtime.

\subsection{Public Suite}

\begin{promptbox}[Intent Classification]
\begin{verbatim}
Classify the utterance intent.
Return JSON with keys: intent,
domain, language.

The "intent" MUST be exactly one of
these 27 labels: [27 intents]
The "domain" MUST be exactly one of
these 11 labels: [11 domains]
The "language" is the ISO 639-1 code
of the utterance (e.g. en).

Dataset: [Dataset]
Utterance: [Utterance]
\end{verbatim}
\end{promptbox}

\begin{promptbox}[Question Answering]
\begin{verbatim}
Answer the question using only the
retrieved content. Return JSON with
keys: answer, evidence, article_ids.
[Retrieved Content]
[Question]
\end{verbatim}
\end{promptbox}

\begin{promptbox}[Dialogue Summarization]
\begin{verbatim}
Summarize the conversation or
meeting. Return JSON with keys:
summary, source_dataset.
[Conversation]
\end{verbatim}
\end{promptbox}

\begin{promptbox}[Chatbot Response]
\begin{verbatim}
Choose the correct tool-use decision
for the user request. Return JSON
with keys: decision, answer,
source_dataset.
Tools: [Tool Specifications]
[User Query]
\end{verbatim}
\end{promptbox}

\subsection{Internal Suite}

Each internal prompt follows the format below, where the
task-specific instruction is placed in the \texttt{[Prompt]} field and the
call transcript in the \texttt{[Transcript]} field:

\begin{promptbox}[Internal prompt format]
\begin{verbatim}
[Prompt]
# Transcript Start #
[Transcript]
# Transcript End #
[Response]
\end{verbatim}
\end{promptbox}

\begin{promptbox}[Action Items]
\begin{verbatim}
Generate a newline separated list of
work, business or service related
TODO tasks that are still not done
at the end of the conversation and
should be completed after the
conversation. [Transcript]
\end{verbatim}
\end{promptbox}

\begin{promptbox}[Call Outcome]
\begin{verbatim}
For the following conversation
transcript, select the best
category from the list provided
below to describe the outcome of
the conversation. Respond with
"Other" if no category applies.
Do not respond with any words
other than the category. The
categories: [List of Categories].
\end{verbatim}
\end{promptbox}

The call-outcome label space is fixed at 30 categories. Examples include: call back, unsuccessful contact, voicemail success, payment / billing, status update, scheduled appointment, cancellation. 

\begin{promptbox}[Meeting Summaries]
\begin{verbatim}
Generate a {Length Type} summary of
the following conversation {Format}
without assessing its quality.
\end{verbatim}
\end{promptbox}
\noindent Here, \texttt{\{Length Type\}} specifies the desired length (very short,
short, long, or descriptive) and \texttt{\{Format\}} the desired format (e.g.\
free text, bullet points, or an array of executive summaries).

\begin{table*}[t]
\centering\small
\setlength{\tabcolsep}{5pt}
\begin{tabular}{lrrrrrrrr}
\toprule
& \multicolumn{4}{c}{Public suite} & \multicolumn{4}{c}{Internal suite} \\
Base & Zero-shot & ST--Full & MT--Full & ST$-$MT & Zero-shot & ST--Full & MT--Full & ST$-$MT \\
\midrule
Qwen3-0.6B & 31.7 & 54.2 & 53.5 & +0.7 & 19.6 & 59.3 & 60.4 & -1.1 \\
Qwen3-1.7B & 43.6 & 56.7 & 56.7 & 0.0 & 31.8 & \edit{64.5} & 58.0 & \edit{+6.5} \\
Qwen3.5-2B & 46.4 & 57.5 & 55.9 & +1.6 & 34.8 & 67.0 & 68.0 & -1.0 \\
Qwen3-4B & 48.7 & 58.0 & 56.0 & +2.0 & 41.8 & 68.7 & 66.8 & +1.9 \\
Qwen3.5-4B & 49.3 & 58.9 & 56.8 & +2.1 & 42.2 & 68.9 & 68.3 & +0.6 \\
Qwen3-8B & 51.6 & 58.6 & 58.1 & +0.5 & 44.7 & 69.1 & 67.5 & +1.6 \\
Qwen3.5-9B & 51.9 & 58.7 & 58.3 & +0.4 & 48.9 & 67.6 & 69.6 & -2.0 \\
Qwen3-14B & 52.1 & 58.8 & 56.0 & +2.8 & 51.2 & 68.7 & 68.2 & +0.5 \\
Qwen3-32B & 52.8 & 59.3 & 58.8 & +0.5 & 53.1 & 70.4 & 68.0 & +2.4 \\
Gemma-3-1B & 39.1 & 55.2 & 55.2 & 0.0 & 25.3 & 64.6 & 62.3 & +2.3 \\
Gemma-3-4B & 51.4 & 57.1 & 54.8 & +2.3 & 41.9 & 68.1 & 67.6 & +0.5 \\
Llama-3.1-8B & 48.4 & 58.0 & 56.4 & +1.6 & 42.3 & 69.2 & 67.8 & +1.4 \\
Mistral-7B-v0.3 & 48.7 & 55.7 & 54.4 & +1.3 & 36.3 & 64.2 & 66.1 & -1.9 \\
\bottomrule
\end{tabular}
\caption{Suite-average results for all models. ST$-$MT denotes the difference between single-task and multi-task full fine-tuning.}
\label{tab:allmodels}
\end{table*}

\begin{promptbox}[Purpose of Call]
\begin{verbatim}
For the conversation below, identify
a single category for the purpose of
the conversation chosen from this
list: [List of Categories]
\end{verbatim}
\end{promptbox}

\new{Purpose of call differs from the other label tasks in that its candidate list is supplied per example rather than fixed: prompts come in several templates, each offering a different shortlist of 8--10 categories drawn from a broader taxonomy of roughly 33 observed in the training data. Examples include: user education, cancellation, support, status inquiry, product support, promotions, account support, maintenance. Because the offered list varies by item, we exclude this task from the output-format-validity analysis in the paper.}

\section{\new{Full Model Coverage}}
\label{app:allmodels}

\newpar{\edit{The main table reports seven representative models from the Qwen3 and Qwen3.5 families spanning 0.6B--9B; Table~\ref{tab:allmodels} extends the comparison to the remaining model sizes and families, providing the internal- and public-suite averages for all thirteen models, together with each model's zero-shot competence on the same suite.}}

\section{Training and Inference Parameters}
\label{hyperparameters}
All training experiments were run across 56 H200 GPUs. The training pipeline is implemented using LLaMA-Factory \citep{zheng2024llamafactory} with DeepSpeed ZeRO-2 ($\leq$8B) / ZeRO-3 (9B) \citep{rajbhandari2020zero}.

For inference, we use vLLM \citep{kwon2023vllm} using tensor-parallel-8 on 8 A100-80GB with the following decoding parameters: temperature $= 0$. 

\section{Model Merging: Effect of the Interpolation Weight $\alpha$}
\label{app:alpha}

\begin{figure}[t]
    \centering
    \includegraphics[width=\linewidth]{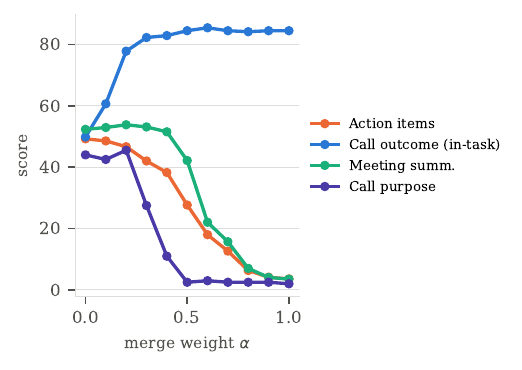}
    \caption{\small{Effect of the interpolation weight $\alpha$ when merging
    the Qwen3.5-9B \texttt{call outcome} specialist with its base model,
    evaluated on all four internal tasks. }}
    \label{fig:alpha-sweep}
\end{figure}

In Section~\ref{sec:merge}, the merge-with-base results primarily use a fixed
interpolation weight $\alpha=0.5$ for consistency. Here we examine
how $\alpha$ trades off in-task specialization against off-task robustness.
Figure~\ref{fig:alpha-sweep} sweeps $\alpha\in[0,1]$ for the Qwen3.5-9B
\texttt{call outcome} specialist merged with its base
($\theta_{\text{merged}}=\alpha\,\theta_{\text{FT}}+(1-\alpha)\,\theta_{\text{base}}$),
evaluating the merged model on all four internal tasks. At $\alpha=0$ the
merged model reduces to the base, while at $\alpha=1$ it recovers the
unmerged specialist.

\new{Two trends are evident. The in-task call-outcome score rises sharply and
saturates early: it reaches $77.8$ by $\alpha=0.2$ and $82.2$ by $\alpha=0.3$,
against $84.4$ for the unmerged specialist, so essentially all of the in-task
benefit is bought by $\alpha\approx0.3$. The three off-task scores instead stay
near their zero-shot values for small $\alpha$ and then collapse: off-task action
items fall from $46.7$ at $\alpha=0.2$ to $3.6$ for the pure specialist, meeting
summaries from $53.8$ to $3.5$, and purpose of call degrades to near zero beyond
$\alpha=0.3$. Purpose of call is the most fragile, losing most of its capability
between $\alpha=0.2$ and $\alpha=0.4$. \edit{For this specialist, $\alpha\in[0.2,0.3]$ provides a useful validation range: at $\alpha=0.3$, the in-task score is within 2.2 points of the specialist, while $\alpha=0.2$ retains more off-task capability. The preferred weight depends on the relative importance of these tasks.}}
We keep $\alpha=0.5$ in the main-text tables for consistency across all merge
configurations, but recommend other $\alpha$ values in practice depending on in-task vs off-task priority. \edit{These results motivate interpolation as an empirical robustness intervention, related to prior work on linear mode connectivity \citep{frankle2020linear} and model soups \citep{wortsman2022model}. We do not measure loss along the interpolation path, so the results do not establish that the specialist and its base lie in the same low-loss region.}

\section{\edit{Training Time and Serving Considerations}}
\label{sec:cost}

\editpar{Table~\ref{tab:costs} reports measured training wall-clock times, rather than serving costs. Each value represents one five-epoch MT-Full run over all four tasks. With the same epochs and no resampling, four ST-Full runs and one MT-Full run process the same total examples; actual compute and wall-clock time also depend on sequence lengths, batching, padding, and hardware utilization. The operational advantage of MT-Full is one training job and checkpoint per suite, without routing among specialist checkpoints. At a fixed model size, per-request inference costs also depend on context length and output length. Maintaining $N$ full specialists can increase model-weight memory approximately $N$-fold, but total serving resources need not scale by $N$ because they depend on traffic, batching, placement, and model loading.}

\begin{table}[t]
\centering\small
\setlength{\tabcolsep}{8pt}
\begin{tabular}{cc}
\toprule
\textbf{Parameters} & \textbf{Training time} \\
\midrule
4B  & \edit{3.1\,h} \\
7B  & \edit{2.1\,h} \\
8B  & \edit{2.1\,h} \\
14B & \edit{1.7\,h} \\
32B & \edit{13.1\,h} \\
\bottomrule
\end{tabular}
\caption{\small{\new{Measured wall-clock time for one five-epoch MT-Full run over four tasks using eight H200 GPUs.}}}
\label{tab:costs}
\end{table}

\end{document}